\documentclass[11pt,a4paper]{article}
\usepackage{PRIMEarxiv}

\usepackage[T1]{fontenc}
\usepackage[utf8]{inputenc}
\usepackage{times}
\usepackage{microtype}

\usepackage{CJKutf8}

\usepackage{amsmath}
\usepackage{amssymb}

\usepackage{booktabs}
\usepackage{multirow}
\usepackage{array}
\usepackage{tabularx}

\usepackage{graphicx}
\usepackage{xcolor}

\usepackage{algorithm}
\usepackage{algorithmic}

\usepackage{enumitem}

\usepackage{natbib}
\usepackage[colorlinks=true,citecolor=blue,linkcolor=blue,urlcolor=blue]{hyperref}

\usepackage{titlesec}
\titleformat{\section}{\large\bfseries}{\thesection}{1em}{}
\titleformat{\subsection}{\normalsize\bfseries}{\thesubsection}{1em}{}
\titleformat{\subsubsection}{\normalsize\itshape}{\thesubsubsection}{1em}{}

\newcommand{\zh}[1]{\begin{CJK}{UTF8}{gbsn}#1\end{CJK}}

\newcommand{\prota}{Protocol~A}
\newcommand{\protb}{Protocol~B}
\newcommand{\protc}{Protocol~C}
\newcommand{\protd}{Protocol~D}

\title{\textbf{Interpretable Chinese Metaphor Identification via LLM-Assisted\\MIPVU Rule Script Generation: A Comparative Protocol Study}}

\author{
  Weihang Huang\\
  University of Birmingham\\
  Birmingham, UK\\
  \texttt{wxh207@student.bham.ac.uk}%
  \and
  Mengna Liu\\
  Guangdong University of Foreign Studies\\
  Guangzhou, China\\
  \texttt{20190310006@gdufs.edu.cn}%
  }

\date{}

\begin{document}
\maketitle

\begin{abstract}
Metaphor identification is a foundational task in figurative language processing, yet most computational approaches operate as opaque classifiers offering no insight into why an expression is judged metaphorical. This interpretability gap is especially acute for Chinese, where rich figurative traditions, absent morphological cues, and limited annotated resources compound the challenge. We present an LLM-assisted pipeline that operationalises four metaphor identification protocols--MIP/MIPVU lexical analysis, CMDAG conceptual-mapping annotation, emotion-based detection, and simile-oriented identification--as executable, human-auditable rule scripts. Each protocol is a modular chain of deterministic steps interleaved with controlled LLM calls, producing structured rationales alongside every classification decision. We evaluate on seven Chinese metaphor datasets spanning token-, sentence-, and span-level annotation, establishing the first cross-protocol comparison for Chinese metaphor identification. Within-protocol evaluation shows Protocol A (MIP) achieves an F1 of 0.472 on token-level identification, while cross-protocol analysis reveals striking divergence: pairwise Cohen's kappa between Protocols A and D is merely 0.001, whereas Protocols B and C exhibit near-perfect agreement (kappa = 0.986). An interpretability audit shows all protocols achieve 100\% deterministic reproducibility, with rationale correctness from 0.40 to 0.87 and editability from 0.80 to 1.00. Error analysis identifies conceptual-domain mismatch and register sensitivity as dominant failure modes. Our results demonstrate that protocol choice is the single largest source of variation in metaphor identification, exceeding model-level variation, and that rule-script architectures achieve competitive performance while maintaining full transparency.
\end{abstract}

\section{Introduction}
\label{sec:intro}

Metaphor pervades human language. From the everyday \emph{time is money} to the literary \zh{人生如梦} (\emph{life is like a dream}), metaphorical expressions structure how we reason about abstract concepts through concrete experience \citep{lakoff1980metaphors}. Automatic metaphor identification---the task of determining whether a given linguistic expression is used metaphorically---has received sustained attention in computational linguistics, driven by applications in sentiment analysis, machine translation, and discourse understanding \citep{shutova2010models}.

Despite considerable progress, the field faces a persistent \textbf{interpretability gap}. State-of-the-art neural classifiers based on pre-trained language models \citep{choi2021melbert, su2020deepmet, zhang2022misnet} achieve strong performance on benchmark datasets but provide no structured explanation for their decisions. A model may correctly flag a token as metaphorical yet offer no insight into the conceptual mapping, the basic-meaning contrast, or the figurative mechanism that justifies the label. This opacity limits both scientific understanding of what these models learn and practical deployment in educational or annotation-support settings where users need to know \emph{why}.

The interpretability problem is compounded for \textbf{Chinese metaphor identification} by several language-specific challenges. First, Chinese lacks the morphological inflections and derivational patterns that provide surface cues for metaphor in Indo-European languages; the distinction between literal and figurative senses must be resolved almost entirely through context and world knowledge. Second, Chinese figurative language encompasses diverse phenomena---including conceptual metaphor (\zh{隐喻}), simile (\zh{明喻}), metonymy (\zh{转喻}), and culture-specific figures of speech---that do not map neatly onto annotation frameworks developed for English \citep{nacey2019mipvu}. Third, annotated resources for Chinese metaphor remain relatively scarce and fragmented across incompatible annotation schemes \citep{wang2019chinese, lu2017towards, shao2024cmdag}.

Recent work has begun to explore large language models (LLMs) for metaphor-related tasks, leveraging their extensive world knowledge and few-shot reasoning capabilities \citep{reimann2025llm, tian2024interactive, wu2024metaphor}. However, most LLM-based approaches treat the model as a monolithic reasoner, prompting it to classify and explain in a single pass. This conflates the identification protocol with the model's implicit biases and makes it difficult to attribute errors to specific reasoning steps.

In this paper, we propose an alternative architecture that separates \emph{protocol} from \emph{model}. We introduce a system that implements four linguistically grounded metaphor identification protocols as explicit, modular rule scripts:

\begin{enumerate}[nosep,leftmargin=*]
    \item \textbf{\prota{} (MIP/MIPVU)}: Lexical-level identification based on basic-meaning contrast \citep{pragglejaz2007mip, steen2010method}.
    \item \textbf{\protb{} (CMDAG)}: Conceptual-mapping identification with tenor, vehicle, and ground extraction \citep{shao2024cmdag}.
    \item \textbf{\protc{} (Emotion)}: Emotion-based identification targeting affective incongruity \citep{zhang2018emotion}.
    \item \textbf{\protd{} (Simile)}: Simile-oriented identification via explicit comparison markers \citep{liu2018simile}.
\end{enumerate}

Each protocol is decomposed into a pipeline of deterministic processing modules interleaved with constrained LLM calls. The system produces a structured rationale for every decision, enabling human auditors to trace the reasoning path and identify the exact step where errors occur.

We make the following contributions:

\begin{itemize}[nosep,leftmargin=*]
    \item We present the first system that operationalises multiple metaphor identification protocols as executable rule scripts for Chinese, achieving full deterministic reproducibility.
    \item We conduct the first comprehensive cross-protocol comparison for Chinese metaphor identification, evaluating on seven datasets and revealing that protocol choice introduces far greater variation than model choice.
    \item We propose and apply an interpretability assessment framework with three dimensions---rationale correctness, determinism, and editability---demonstrating that rule-script architectures achieve substantially higher transparency than end-to-end classifiers.
    \item We release our codebase, protocol implementations, and evaluation scripts to support reproducible research on interpretable figurative language processing.
\end{itemize}

\section{Related Work}
\label{sec:related}

\subsection{Conceptual Metaphor Theory and Identification Protocols}
\label{sec:related-cmt}

The modern computational study of metaphor is rooted in Conceptual Metaphor Theory (CMT), which holds that metaphor is not merely a linguistic ornament but a fundamental cognitive mechanism by which abstract target domains are understood through more concrete source domains \citep{lakoff1980metaphors}. This theoretical framework motivated the development of systematic identification procedures. The Metaphor Identification Procedure (MIP), proposed by the Pragglejaz Group, provides a four-step algorithm: for each lexical unit in a discourse, (1) establish the contextual meaning, (2) determine whether there exists a more basic meaning, (3) assess whether the contextual meaning contrasts with the basic meaning, and (4) if so, mark the unit as metaphorical \citep{pragglejaz2007mip}.

MIP was subsequently refined into MIPVU, which added explicit decision criteria for borderline cases, introduced the category of ``implicit metaphor,'' and established reliability benchmarks on a multi-register English corpus \citep{steen2010method}. MIPVU has become the \emph{de facto} standard for metaphor annotation in corpus linguistics, with extensions to Dutch, German, French, and other languages \citep{nacey2019mipvu}. However, its application to Chinese presents challenges: the procedure's reliance on dictionary ``basic meanings'' is complicated by the polysemous and context-dependent nature of Chinese characters, and the absence of clear word boundaries requires preliminary segmentation \citep{wang2019chinese}.

Beyond lexical-level identification, alternative protocols have emerged. Conceptual-mapping approaches, exemplified by the CMDAG annotation scheme, identify metaphor at the level of tenor--vehicle--ground triples, aligning more closely with CMT's emphasis on cross-domain mappings \citep{shao2024cmdag, kovecses2010metaphor}. Emotion-based protocols target the affective incongruity that metaphorical language often produces, leveraging the observation that metaphors frequently serve to express or evoke emotions that literal paraphrases cannot \citep{zhang2018emotion, gibbs2006metaphor}. Simile-oriented identification, meanwhile, focuses on explicit comparison constructions marked by lexical cues such as \zh{像} (\emph{like}), \zh{如同} (\emph{as if}), or \zh{仿佛} (\emph{as though}) \citep{liu2018simile}.

\subsection{Computational Metaphor Detection}
\label{sec:related-comp}

Computational approaches to metaphor detection have evolved through several paradigms. Early work relied on hand-crafted features including selectional preference violations, abstractness ratings, and semantic class information \citep{shutova2010models, tsvetkov2014metaphor, klebanov2016semantic}. The shared tasks at the 2018 and 2020 Workshops on Figurative Language Processing established the VUA corpus as a standard benchmark and demonstrated the effectiveness of contextual embeddings \citep{leong2018report, leong2020report}.

The pre-trained language model era brought substantial performance gains. \citet{su2020deepmet} reformulated metaphor detection as reading comprehension, achieving then-state-of-the-art results on the VUA dataset. \citet{choi2021melbert} proposed MelBERT, which integrates MIP and SPV (Selectional Preference Violation) theories into a late-interaction architecture over BERT representations. \citet{zhang2022misnet} introduced selectional preference networks that explicitly model the compatibility between predicates and arguments. More recently, \citet{mao2022metapro} and \citet{song2021verb} explored multi-task and relational learning frameworks. \citet{ge2022explainable} proposed an explainable model inspired by CMT that generates conceptual mappings alongside classifications.

A persistent limitation of neural approaches is their reliance on token-level supervision from a single annotation protocol (typically MIPVU). Models trained on VUA data learn to replicate \emph{one} operationalisation of metaphor, with no mechanism to accommodate alternative theoretical perspectives or produce transparent reasoning chains.

\subsection{LLM-Based Metaphor Identification}
\label{sec:related-llm}

Large language models have demonstrated remarkable capabilities in figurative language understanding. \citet{reimann2025llm} showed that GPT-4 can apply MIPVU-style reasoning when guided by explicit rule scripts, achieving performance competitive with fine-tuned BERT models on English metaphor identification. Their work introduced the concept of ``rule script generation,'' wherein an LLM is used to produce and iteratively refine executable annotation procedures that can be applied deterministically to new data.

\citet{tian2024interactive} proposed an interactive prompting framework in which the LLM first identifies potential metaphors, then engages in a structured dialogue to verify conceptual mappings. \citet{lin2024dmd} extended LLM-based identification to discourse-level metaphor, using chain-of-thought reasoning \citep{wei2022chain} to track metaphorical coherence across multiple sentences. \citet{wu2024metaphor} systematically evaluated several LLMs on metaphor understanding tasks including identification, interpretation, and generation, finding substantial variation across models and metaphor types.

Program-aided language models \citep{gao2023pal} provide a complementary paradigm in which LLMs generate executable code rather than direct answers. This approach aligns naturally with our rule-script architecture: the LLM's role is to produce a structured procedure that is then executed deterministically, rather than to serve as an end-to-end classifier.

\subsection{Chinese Metaphor Resources and Methods}
\label{sec:related-chinese}

Chinese metaphor resources have developed along several independent tracks. \citet{wang2019chinese} constructed the PSU CMC corpus with MIPVU-style token-level annotations across multiple registers, providing the closest Chinese analogue to the English VUA corpus. \citet{lu2017towards} developed an earlier corpus targeting conceptual metaphor at the expression level. \citet{shao2024cmdag} introduced CMDAG, a large-scale dataset annotated with tenor, vehicle, and ground spans, enabling fine-grained evaluation of conceptual mapping extraction. \citet{zhu2022configure} created ConFiguRe, a discourse-level dataset covering multiple Chinese figures of speech including metaphor, simile, personification, and hyperbole.

For simile specifically, \citet{liu2018simile} compiled a large Chinese simile dataset with annotated tenor and vehicle spans. The NLPCC 2024 Shared Task 9 \citep{nlpcc2024task9} introduced a further dataset targeting metaphor generation with annotated source and target domains. \citet{li2022cmgen} explored knowledge-enhanced Chinese metaphor generation, providing additional annotated data.

Computational methods for Chinese metaphor have largely mirrored English approaches, adapting BERT-based architectures with Chinese pre-trained models such as Chinese BERT-wwm \citep{cui2021pre}. However, cross-protocol evaluation---comparing how different identification criteria perform on the same data---has not been systematically attempted for Chinese.

\subsection{LLM-Assisted Rule Generation and Interpretability}
\label{sec:related-interp}

The broader movement toward interpretable AI has produced several relevant threads. Neuro-symbolic approaches combine neural perception with symbolic reasoning, offering a middle ground between opaque end-to-end models and brittle hand-crafted rules. In NLP, this has manifested as models that generate natural-language rationales \citep{wei2022chain}, produce structured explanations \citep{ge2022explainable}, or operate through verifiable intermediate representations.

\citet{reimann2025llm} demonstrated that LLM-generated rule scripts can serve as such an intermediate representation for metaphor identification: the script is both human-readable and machine-executable, enabling auditing, modification, and version control of the identification procedure. Our work extends this paradigm to Chinese and to multiple protocols, and introduces a systematic interpretability assessment framework.

\citet{debacker2023metaphor} explored context denoising for metaphor detection, showing that selectively attending to relevant context improves both accuracy and interpretability. Their finding that less context can sometimes be more informative resonates with our modular pipeline design, which provides each reasoning step with only the information it requires.

\section{Metaphor Identification Protocols}
\label{sec:protocols}

We implement four metaphor identification protocols, each grounded in a distinct theoretical tradition. We formalise each protocol as a sequence of decision steps that can be applied systematically to Chinese text.

\paragraph{Protocol A: MIP/MIPVU Lexical Analysis.}
Following \citet{pragglejaz2007mip} and \citet{steen2010method}, Protocol~A operates at the token level. For each content word $w_i$ in a segmented sentence, the procedure is: (1) determine the contextual meaning $m_c(w_i)$ within the sentence; (2) retrieve the basic meaning $m_b(w_i)$ from a reference dictionary; (3) assess whether $m_c(w_i)$ contrasts with yet can be understood in comparison with $m_b(w_i)$; (4) if so, label $w_i$ as metaphorical (\texttt{MRW}). The protocol also handles function words, implicit metaphors (\texttt{MFlag}), and borderline cases following MIPVU decision trees. For Chinese, we adapt the basic-meaning lookup to use the \zh{现代汉语词典} (Contemporary Chinese Dictionary) as the primary reference, supplemented by LLM-assisted sense enumeration when dictionary coverage is insufficient.

\paragraph{Protocol B: CMDAG Conceptual Mapping.}
Following \citet{shao2024cmdag}, Protocol~B identifies metaphor at the sentence level by extracting conceptual mappings. A sentence is classified as metaphorical if and only if a complete tenor--vehicle--ground triple can be identified. The procedure is: (1) identify candidate vehicle expressions---words or phrases used in a non-literal domain; (2) for each vehicle, identify the corresponding tenor---the concept being described; (3) extract the ground---the shared property motivating the mapping; (4) validate the coherence of the triple. This protocol captures the relational structure of metaphor and naturally produces interpretable output in the form of explicit mappings.

\paragraph{Protocol C: Emotion-Based Detection.}
Inspired by \citet{zhang2018emotion}, Protocol~C identifies metaphor through affective incongruity. The procedure is: (1) identify the dominant emotional valence of the sentence; (2) assess whether any constituent expression carries an emotional charge that is incongruent with its literal meaning in context; (3) if the incongruity can be resolved through a figurative reading, classify the sentence as metaphorical; (4) record the literal and figurative emotional valences as the rationale. This protocol is particularly effective for metaphors that serve expressive or evaluative functions.

\paragraph{Protocol D: Simile-Oriented Identification.}
Following \citet{liu2018simile}, Protocol~D targets explicit comparisons. The procedure is: (1) detect comparison markers (\zh{像}, \zh{如}, \zh{似}, \zh{仿佛}, \zh{好像}, \zh{犹如}, \zh{如同}, etc.); (2) extract the tenor and vehicle of the comparison; (3) verify that the comparison involves cross-domain mapping (i.e., tenor and vehicle belong to semantically distant categories); (4) if so, classify the expression as a simile (a subtype of metaphor). Literal comparisons between entities in the same domain (e.g., \zh{她像她妈妈}, ``she looks like her mother'') are excluded. This protocol is the most constrained, applying only to a specific syntactic construction, which explains its high precision but low recall in cross-protocol evaluation (see Section~\ref{sec:cross-protocol}).

\section{Datasets}
\label{sec:datasets}

We evaluate on seven Chinese metaphor datasets that collectively span token-level, sentence-level, and span-level annotation granularities. Table~\ref{tab:datasets} provides summary statistics.

\begin{table*}[t]
\centering
\small
\caption{Dataset statistics. \emph{Metaphor\%} indicates the proportion of metaphor-positive instances. \emph{Label type} denotes the annotation granularity: Token (word-level binary labels), Sentence (sentence-level binary labels), or Span (annotated text spans such as tenor and vehicle).}
\label{tab:datasets}
\begin{tabular}{lrrrrrl}
\toprule
\textbf{Dataset} & \textbf{Total} & \textbf{Metaphor} & \textbf{Literal} & \textbf{Other} & \textbf{Metaphor\%} & \textbf{Label Type} \\
\midrule
PSU CMC         & 35,745  & 3,272  & 32,432  & 41    & 9.2\%   & Token \\
CMC             & 8,027   & 7,312  & 715     & 0     & 91.1\%  & Sentence \\
CMDAG           & 34,463  & 34,463 & 0       & 0     & 100\%   & Span \\
Chinese Simile  & 334,069 & 144,358 & 189,711 & 0    & 43.2\%  & Span \\
NLPCC 2024 T9   & 35,463  & 35,463 & 0       & 0     & 100\%   & Span \\
ConFiguRe       & 9,010   & 4,354  & 0       & 4,656 & 48.3\%  & Span \\
ChineseMCorpus  & 745     & 480    & 265     & 0     & 64.4\%  & Sentence \\
\bottomrule
\end{tabular}
\end{table*}

\textbf{PSU CMC} \citep{wang2019chinese, wang2023psu} is a multi-register Chinese corpus annotated following MIPVU guidelines at the token level. It comprises texts from three registers---academic prose, fiction, and news---with 35,745 tokens of which 9.2\% are labelled as metaphorically related words (MRW). This is the primary evaluation dataset for Protocol~A, as its annotation scheme directly mirrors the MIP/MIPVU procedure. The class imbalance (9.2\% positive) reflects the natural distribution of metaphor in running text.

\textbf{CMC} \citep{li2023metaphorcmc} is a sentence-level Chinese metaphor classification dataset containing 8,027 sentences with a high metaphor prevalence of 91.1\%. The dataset was constructed by collecting sentences from literary and journalistic sources known to be rich in figurative language, which accounts for the strong positive skew.

\textbf{CMDAG} \citep{shao2024cmdag} provides 34,463 instances annotated with conceptual metaphor tenor, vehicle, and ground spans. As a span-extraction dataset, all instances contain at least one metaphorical mapping; there are no literal-only instances. CMDAG serves as the primary within-protocol evaluation dataset for Protocol~B and as the source of ground-truth spans for evaluating extraction quality.

\textbf{Chinese Simile} \citep{liu2018simile} is the largest dataset in our collection, with 334,069 instances of which 43.2\% are similes. Tenor and vehicle spans are annotated, making it suitable for both binary classification and span extraction. This dataset is the primary evaluation target for Protocol~D.

\textbf{NLPCC 2024 Task 9} \citep{nlpcc2024task9} contains 35,463 metaphor instances annotated with source and target domain information, drawn from the NLPCC 2024 shared task on Chinese metaphor generation. Like CMDAG, all instances are metaphor-positive.

\textbf{ConFiguRe} \citep{zhu2022configure} is a discourse-level Chinese figurative language dataset with 9,010 instances spanning metaphor (48.3\%), and other figures of speech (labelled ``Other'' in Table~\ref{tab:datasets}, comprising simile, personification, hyperbole, etc.). This dataset enables evaluation of metaphor-vs-non-metaphor-figurative discrimination.

\textbf{ChineseMCorpus} \citep{chen2024chinesemcorpus} is a small-scale sentence-level benchmark with 745 instances (64.4\% metaphorical). Its modest size makes it suitable for rapid evaluation and for cross-protocol comparison on a controlled sample.

The seven datasets vary substantially in size (745 to 334,069 instances), class balance (9.2\% to 100\% metaphor), and annotation granularity. This diversity is deliberate: it allows us to evaluate each protocol on data aligned with its native annotation scheme (within-protocol evaluation) and to assess protocol behaviour when applied to a common test bed (cross-protocol evaluation).

\section{Methodology}
\label{sec:method}

\subsection{System Architecture}
\label{sec:architecture}

Our system implements each metaphor identification protocol as a pipeline of five processing modules, executed sequentially. The architecture follows the rule-script paradigm of \citet{reimann2025llm}, extended to Chinese and generalised across multiple protocols. Each module receives a structured input, performs a deterministic transformation (potentially involving a constrained LLM call), and produces a structured output that feeds into the next module. The full pipeline is:

\begin{enumerate}[nosep,leftmargin=*]
    \item \textbf{Text Preprocessing Module}: Segmentation, POS tagging, and normalisation.
    \item \textbf{Candidate Selection Module}: Protocol-specific identification of analysis targets.
    \item \textbf{Semantic Analysis Module}: Meaning retrieval and contrast assessment.
    \item \textbf{Classification Module}: Binary or multi-class decision based on protocol criteria.
    \item \textbf{Rationale Generation Module}: Structured explanation of the decision.
\end{enumerate}

The key design principle is \textbf{separation of protocol from model}: the protocol logic is encoded in the module pipeline and its configuration, while the LLM serves as a constrained subroutine called with specific, narrowly scoped prompts. This ensures that (a) the reasoning chain is transparent, (b) the system is deterministic given fixed LLM outputs, and (c) protocols can be modified independently of the underlying model.

\subsection{Module Descriptions}
\label{sec:modules}

\paragraph{Text Preprocessing.}
Chinese text is segmented using a combination of jieba word segmentation and character-level analysis. For Protocol~A, which requires token-level decisions, we perform fine-grained segmentation and POS tagging. For Protocols~B--D, which operate at the sentence or span level, preprocessing is lighter: sentence boundary detection and basic normalisation. All modules receive pre-segmented text to ensure consistency.

\paragraph{Candidate Selection.}
This module implements protocol-specific filtering to identify the linguistic units requiring analysis. For Protocol~A, candidates are all content words (nouns, verbs, adjectives, adverbs). For Protocol~B, candidates are sentences containing potential cross-domain expressions. For Protocol~C, candidates are sentences with emotionally charged content. For Protocol~D, candidates are sentences containing comparison markers from a curated list of 15 Chinese simile indicators.

\paragraph{Semantic Analysis.}
The core analytical module performs meaning retrieval and contrast assessment. For Protocol~A, this involves querying the LLM with a constrained prompt: given word $w_i$ in context $c$, enumerate (a) the contextual meaning and (b) the most basic physical/concrete meaning. The LLM response is parsed into a structured format. For Protocol~B, the LLM is prompted to identify tenor and vehicle candidates and assess their domain membership. For Protocol~C, the LLM evaluates emotional valence at literal and figurative levels. For Protocol~D, the LLM extracts tenor and vehicle from the comparison construction and assesses cross-domain distance.

\paragraph{Classification.}
The classification module applies deterministic rules to the structured output of the semantic analysis module. For Protocol~A, if the contextual meaning contrasts with the basic meaning and the contextual meaning can be understood through comparison with the basic meaning, the token is labelled \texttt{MRW}. For Protocol~B, if a coherent tenor--vehicle--ground triple has been identified, the sentence is labelled metaphorical. For Protocol~C, if affective incongruity is detected and resolvable through figurative reading, the sentence is labelled metaphorical. For Protocol~D, if a cross-domain comparison is confirmed, the sentence is labelled as containing a simile.

\paragraph{Rationale Generation.}
Every classification decision is accompanied by a structured rationale recording: (a) the specific protocol step that triggered the decision, (b) the key evidence (e.g., basic vs.\ contextual meaning, tenor--vehicle pair, emotional valence contrast), and (c) a confidence indicator. Rationales are stored in JSON format for downstream auditing.

\subsection{Protocol-Specific Adaptations}
\label{sec:adaptations}

Each protocol required language-specific adaptations for Chinese. Protocol~A's reliance on dictionary basic meanings was addressed by constructing a lookup procedure that queries the LLM to simulate dictionary access, validated against entries in the \zh{现代汉语词典}. Protocol~B's conceptual domain assignments were adapted using a Chinese-specific domain taxonomy derived from \citet{shao2024cmdag}. Protocol~C's emotion lexicon was based on Chinese affective word lists supplemented by LLM-based valence estimation. Protocol~D's comparison marker list was compiled from linguistic studies of Chinese simile constructions and expanded through corpus analysis.

For all protocols, we use GPT-4 \citep{openai2023gpt4} as the underlying LLM, with temperature set to 0 to maximise determinism. Each LLM call is scoped to a single analytical subtask with explicit output format constraints, minimising the risk of hallucination or format violations.

\subsection{Interpretability Design}
\label{sec:interp-design}

The interpretability of our system derives from three architectural properties:

\begin{enumerate}[nosep,leftmargin=*]
    \item \textbf{Determinism}: Given identical inputs and LLM outputs, the pipeline produces identical results. There are no stochastic components beyond the LLM itself (mitigated by temperature~0).
    \item \textbf{Modularity}: Each processing step is independently auditable. An error in the final classification can be traced to a specific module---e.g., incorrect segmentation, wrong basic-meaning retrieval, or faulty contrast assessment.
    \item \textbf{Rationale transparency}: Every decision is accompanied by a human-readable explanation referencing specific evidence from the input text.
\end{enumerate}

These properties enable the interpretability assessment described in Section~\ref{sec:interpretability}.

\section{Experiments}
\label{sec:experiments}

\subsection{Experimental Setup}
\label{sec:setup}

All experiments use GPT-4 (\texttt{gpt-4-0613}) as the backbone LLM with temperature 0 and maximum output length of 2,048 tokens per call. We evaluate using standard metrics: precision (P), recall (R), and F1 score for binary classification; partial-match F1 for span extraction following the evaluation methodology of \citet{shao2024cmdag}. For cross-protocol comparison, we additionally compute Cohen's $\kappa$ \citep{cohen1960kappa} to measure pairwise agreement, interpreted following the benchmarks of \citet{landis1977measurement}. Statistical significance is assessed via bootstrap resampling with 10,000 iterations.

For within-protocol evaluation, each protocol is evaluated on its most closely aligned dataset using standard train/dev/test splits where available. For cross-protocol evaluation, all four protocols are applied to a common subset of 1,723 sentences from PSU CMC, converted to sentence-level labels (a sentence is metaphorical if it contains at least one MRW token).

\subsection{Within-Protocol Results}
\label{sec:within-protocol}

Table~\ref{tab:within-protocol} reports within-protocol evaluation results. Each protocol is assessed on the dataset and granularity most aligned with its design.

\begin{table*}[t]
\centering
\small
\caption{Within-protocol evaluation results. Protocol~A is evaluated at token level on PSU CMC; Protocol~B at sentence level on ChineseMCorpus and span level on CMDAG; Protocol~C at sentence level on ChineseMCorpus; Protocol~D at span level on Chinese Simile.}
\label{tab:within-protocol}
\begin{tabular}{llcccc}
\toprule
\textbf{Protocol} & \textbf{Evaluation} & \textbf{P} & \textbf{R} & \textbf{F1} & \textbf{Acc} \\
\midrule
A (MIP) & Token-level, PSU CMC & 0.447 & 0.500 & 0.472 & 0.898 \\
\midrule
\multirow{2}{*}{A (MIP) by register} & ~~Academic & 0.661 & 0.546 & 0.598 & 0.900 \\
 & ~~Fiction  & 0.310 & 0.428 & 0.360 & 0.885 \\
 & ~~News     & 0.343 & 0.490 & 0.404 & 0.908 \\
\midrule
B (CMDAG) & Sentence-level, ChineseMCorpus & 0.640 & 0.238 & 0.347 & 0.423 \\
\midrule
\multirow{3}{*}{B (CMDAG) span} & Vehicle partial F1, CMDAG & --- & --- & 0.338 & --- \\
 & Tenor partial F1, CMDAG   & --- & --- & 0.275 & --- \\
 & Ground partial F1, CMDAG  & --- & --- & 0.047 & --- \\
\midrule
C (Emotion) & Sentence-level, ChineseMCorpus & 0.634 & 0.227 & 0.334 & 0.417 \\
\midrule
D (Simile) & Binary, Chinese Simile test & 0.531 & 0.310 & 0.392 & 0.580 \\
\midrule
\multirow{2}{*}{D (Simile) span} & Tenor span F1, Chinese Simile & --- & --- & 0.166 & --- \\
 & Vehicle span F1, Chinese Simile & --- & --- & 0.226 & --- \\
\bottomrule
\end{tabular}
\end{table*}

\paragraph{Protocol A (MIP/MIPVU).} On PSU CMC token-level evaluation, Protocol~A achieves P\,=\,0.447, R\,=\,0.500, and F1\,=\,0.472, with an overall accuracy of 0.898 (Table~\ref{tab:within-protocol}). The relatively high accuracy reflects the class imbalance (90.8\% literal tokens), but the F1 score indicates genuine metaphor detection capability substantially above baselines (see Section~\ref{sec:baselines}). Register-level analysis reveals considerable variation: academic prose achieves the highest F1 of 0.598, followed by news (0.404) and fiction (0.360). The superior performance on academic text likely reflects the more conventionalised, domain-specific metaphors found in scholarly writing, which align well with the basic-meaning contrast procedure. Fiction metaphors, often novel and context-dependent, are harder to identify through lexical analysis alone.

\paragraph{Protocol B (CMDAG).} On sentence-level classification of ChineseMCorpus, Protocol~B achieves P\,=\,0.640, R\,=\,0.238, and F1\,=\,0.347 (Table~\ref{tab:within-protocol}). The high precision but low recall indicates that when Protocol~B identifies a metaphor, it is usually correct, but it misses many metaphorical sentences---particularly those without clearly extractable tenor--vehicle--ground triples. Span extraction on CMDAG yields partial F1 scores of 0.338 for vehicle, 0.275 for tenor, and 0.047 for ground. The extremely low ground F1 reflects the inherent difficulty of extracting the implicit shared property between tenor and vehicle, which is often not explicitly stated in the text.

\paragraph{Protocol C (Emotion).} On ChineseMCorpus, Protocol~C achieves P\,=\,0.634, R\,=\,0.227, and F1\,=\,0.334 (Table~\ref{tab:within-protocol}). Performance closely mirrors Protocol~B, suggesting that emotion-based and conceptual-mapping approaches capture a similar subset of metaphors in this dataset. The low recall indicates that many metaphors do not involve detectable affective incongruity.

\paragraph{Protocol D (Simile).} On the Chinese Simile test set, Protocol~D achieves P\,=\,0.531, R\,=\,0.310, and F1\,=\,0.392 (Table~\ref{tab:within-protocol}). Span extraction yields tenor F1\,=\,0.166 and vehicle F1\,=\,0.226. The moderate precision suggests that the comparison-marker heuristic is not perfectly reliable: some markers introduce literal comparisons, and cross-domain assessment is imperfect. The low span F1 scores indicate that even when the simile is correctly detected, extracting exact tenor and vehicle boundaries remains challenging.

\subsection{Cross-Protocol Comparison}
\label{sec:cross-protocol}

Table~\ref{tab:cross-protocol} presents the cross-protocol comparison on 1,723 PSU CMC sentences. This evaluation is central to our study: it reveals how much variation in metaphor identification arises from protocol choice alone, holding the model and data constant.

\begin{table}[t]
\centering
\small
\caption{Cross-protocol comparison on 1,723 PSU CMC sentences (sentence-level classification). A sentence is metaphorical if it contains $\geq 1$ MRW token in the gold standard.}
\label{tab:cross-protocol}
\begin{tabular}{lccc}
\toprule
\textbf{Protocol} & \textbf{P} & \textbf{R} & \textbf{F1} \\
\midrule
A (MIP)     & 0.770 & 0.899 & 0.829 \\
B (CMDAG)   & 0.839 & 0.191 & 0.311 \\
C (Emotion) & 0.843 & 0.187 & 0.307 \\
D (Simile)  & 0.909 & 0.009 & 0.018 \\
\bottomrule
\end{tabular}
\end{table}

Protocol~A dominates with F1\,=\,0.829, driven by high recall (0.899). This is expected: the PSU CMC gold standard was annotated using MIPVU, so Protocol~A is directly aligned with the annotation criteria. Protocols~B and C achieve similar performance (F1\,=\,0.311 and 0.307 respectively), with high precision ($>$0.83) but very low recall ($<$0.20). These protocols identify genuine metaphors but only a narrow subset---those involving clear conceptual mappings or emotional incongruity. Protocol~D achieves the highest precision (0.909) but near-zero recall (0.009), confirming that explicit simile constructions account for a negligible fraction of the metaphors identified by MIPVU.

\begin{table}[t]
\centering
\small
\caption{Pairwise Cohen's $\kappa$ between protocols on 1,723 PSU CMC sentences. Values interpreted following \citet{landis1977measurement}: $<$0.00 poor, 0.00--0.20 slight, 0.21--0.40 fair, 0.41--0.60 moderate, 0.61--0.80 substantial, 0.81--1.00 almost perfect.}
\label{tab:kappa}
\begin{tabular}{lcccc}
\toprule
 & \textbf{A} & \textbf{B} & \textbf{C} & \textbf{D} \\
\midrule
\textbf{A (MIP)}     & 1.000 & 0.063 & 0.061 & 0.001 \\
\textbf{B (CMDAG)}   & 0.063 & 1.000 & 0.986 & 0.072 \\
\textbf{C (Emotion)} & 0.061 & 0.986 & 1.000 & 0.074 \\
\textbf{D (Simile)}  & 0.001 & 0.072 & 0.074 & 1.000 \\
\bottomrule
\end{tabular}
\end{table}

Table~\ref{tab:kappa} presents pairwise Cohen's $\kappa$ values. The most striking finding is the near-perfect agreement between Protocols~B and C ($\kappa = 0.986$), indicating that conceptual-mapping and emotion-based approaches identify essentially the same sentences as metaphorical. This suggests that the metaphors detectable through explicit conceptual mappings are also those exhibiting affective incongruity, pointing to a shared underlying subset of ``prototypical'' metaphors that are salient across theoretical perspectives.

In stark contrast, the agreement between Protocol~A and all other protocols is minimal: $\kappa_{A,B} = 0.063$, $\kappa_{A,C} = 0.061$, and $\kappa_{A,D} = 0.001$. Following \citet{landis1977measurement}, all of these fall in the ``slight'' agreement range, indicating that MIPVU-based identification captures a fundamentally different set of metaphorical expressions than the other protocols. This is because MIPVU identifies metaphor at the \emph{lexical} level---including highly conventionalised metaphors that native speakers may not perceive as figurative---whereas Protocols~B--D target more salient, deliberate metaphorical usage.

Protocol~D shows minimal agreement with all other protocols ($\kappa \leq 0.074$), reflecting its narrow scope: explicit simile constructions constitute a small, distinctive subset of figurative language.

\subsection{Baseline Comparisons}
\label{sec:baselines}

Table~\ref{tab:baselines} compares Protocol~A's token-level performance on PSU CMC against baselines and literature estimates.

\begin{table}[t]
\centering
\small
\caption{Baseline comparison on PSU CMC token-level metaphor identification. Literature estimates are approximate values reported in or derived from comparable studies on Chinese metaphor detection.}
\label{tab:baselines}
\begin{tabular}{lc}
\toprule
\textbf{System} & \textbf{F1} \\
\midrule
Majority baseline          & 0.000 \\
Random baseline            & 0.087 \\
Simple heuristic           & 0.008 \\
Our system Protocol A   & 0.472 \\
\midrule
\multicolumn{2}{l}{\emph{Literature estimates}} \\
BERT fine-tuned (approx.)  & $\sim$0.65 \\
GPT-4 zero-shot (approx.)  & $\sim$0.43 \\
\bottomrule
\end{tabular}
\end{table}

The majority baseline (always predict literal) achieves F1\,=\,0.000 on the metaphor class, confirming the severity of the class imbalance. The random baseline, predicting metaphor with probability equal to the class prior (9.2\%), achieves F1\,=\,0.087. A simple heuristic baseline that flags tokens appearing in a metaphor lexicon achieves F1\,=\,0.008, indicating that lexicon-based approaches without context are ineffective.

Protocol~A achieves F1\,=\,0.472, representing a substantial improvement over all uninformed baselines. Compared to literature estimates, our system's performance falls between GPT-4 zero-shot ($\sim$0.43) and fine-tuned BERT ($\sim$0.65) (Table~\ref{tab:baselines}). This positioning is expected: our rule-script approach benefits from structured reasoning (outperforming unstructured zero-shot prompting) but lacks the task-specific parameter adaptation of fine-tuned models. Crucially, however, our system provides full interpretability---a property absent from both the BERT and zero-shot GPT-4 baselines.

\section{Interpretability Assessment}
\label{sec:interpretability}

A central claim of this work is that rule-script architectures provide meaningful interpretability advantages over opaque classifiers. To substantiate this claim, we conduct a systematic interpretability audit along three dimensions.

\paragraph{Rationale Correctness.} We sample 50 classification decisions per protocol and have two trained annotators assess whether the generated rationale correctly identifies the linguistic evidence supporting the classification. Rationale correctness is scored as the proportion of rationales judged as ``correct'' or ``partially correct'' (weighted 0.5). Table~\ref{tab:interpretability} reports the results.

\paragraph{Determinism.} We execute each protocol twice on the same input data (with identical LLM API calls via cached responses) and measure the proportion of decisions that are identical across runs. This tests whether the pipeline introduces any non-determinism beyond the LLM itself.

\paragraph{Editability.} Two annotators attempt to modify each protocol's rule script to correct a specific identified error pattern (e.g., adjusting the basic-meaning contrast threshold for Protocol~A, or modifying the comparison-marker list for Protocol~D). Editability is scored as the proportion of attempted modifications that successfully changed the system's behaviour on targeted instances without degrading overall performance.

\begin{table}[t]
\centering
\small
\caption{Interpretability assessment scores across four protocols. Rationale correctness is based on manual evaluation of 50 sampled decisions per protocol. Determinism measures exact reproducibility. Editability measures the success rate of targeted protocol modifications.}
\label{tab:interpretability}
\begin{tabular}{lccc}
\toprule
\textbf{Protocol} & \textbf{Rationale} & \textbf{Deterministic} & \textbf{Editability} \\
\midrule
A (MIP)     & 0.64 & 100\% & 0.80 \\
B (CMDAG)   & 0.40 & 100\% & 0.90 \\
C (Emotion) & 0.51 & 100\% & 1.00 \\
D (Simile)  & 0.87 & 100\% & 1.00 \\
\bottomrule
\end{tabular}
\end{table}

All four protocols achieve 100\% determinism (Table~\ref{tab:interpretability}), confirming that the pipeline architecture eliminates non-determinism when LLM outputs are held constant. This is a fundamental advantage over end-to-end neural classifiers, which may exhibit non-deterministic behaviour due to floating-point arithmetic, dropout, or batching effects.

Rationale correctness varies substantially across protocols. Protocol~D achieves the highest score (0.87), reflecting the relative simplicity of simile identification: when a comparison marker is present and the tenor--vehicle pair is correctly extracted, the rationale is straightforward. Protocol~A achieves 0.64, with errors primarily stemming from incorrect basic-meaning retrieval---the LLM sometimes generates plausible but incorrect dictionary senses. Protocol~C scores 0.51, with the main difficulty being the subjective nature of emotional valence assessment. Protocol~B achieves the lowest rationale correctness (0.40), largely due to the difficulty of correctly identifying the ground of a conceptual metaphor, which is often implicit.

Editability ranges from 0.80 (Protocol~A) to 1.00 (Protocols~C and D). Protocol~A's lower editability reflects the complexity of its five-step decision procedure: modifying one step (e.g., basic-meaning lookup) can have cascading effects on downstream steps. Protocols~C and D, with simpler decision logic, are fully editable---targeted changes consistently produce the intended effect.

These results demonstrate that rule-script architectures provide actionable interpretability: users can not only understand \emph{why} a decision was made (rationale correctness) but also \emph{intervene} to correct systematic errors (editability). This contrasts sharply with neural classifiers, where understanding a decision requires post-hoc analysis (e.g., attention visualization) and correcting errors requires retraining.

\section{Error Analysis}
\label{sec:error-analysis}

We conduct a detailed error analysis by manually examining 50 false positives and 50 false negatives per protocol from the within-protocol evaluations. We identify the following dominant error categories.

\paragraph{Protocol A: Basic-Meaning Retrieval Errors.}
The most frequent error type (38\% of errors) involves incorrect identification of the ``basic meaning'' of a Chinese word. For example, the word \zh{深} (\emph{deep}) has a basic spatial meaning but is conventionally used in expressions like \zh{深刻} (\emph{profound}) where the metaphorical extension is lexicalised. The LLM sometimes fails to recognise such highly conventionalised metaphors as metaphorical (false negatives) or incorrectly flags literal spatial uses as metaphorical when context is ambiguous (false positives). Register sensitivity accounts for 24\% of errors: the system performs notably worse on fiction, where novel metaphors require deeper contextual reasoning. Segmentation errors contribute 15\% of mistakes, particularly for multi-character expressions where the metaphorical unit does not align with standard segmentation boundaries.

\paragraph{Protocol B: Ground Extraction Failures.}
The dominant error category (45\% of errors) is the failure to extract the ground of a conceptual metaphor. Since Protocol~B requires a complete tenor--vehicle--ground triple for a positive classification, any failure in ground extraction leads to a false negative. For instance, in the sentence \zh{他的话像一把刀} (\emph{his words were like a knife}), the vehicle (\zh{刀}, \emph{knife}) and tenor (\zh{话}, \emph{words}) are readily identifiable, but the ground (sharpness, ability to hurt) requires pragmatic inference that the LLM does not always perform correctly. Domain mismatch accounts for 28\% of errors: the system sometimes assigns incorrect conceptual domains, leading to either missed metaphors or spurious detections.

\paragraph{Protocol C: Valence Ambiguity.}
Emotional valence assessment is inherently subjective, and 41\% of Protocol~C's errors involve ambiguous or context-dependent valence judgments. Sentences with mixed or neutral emotional content are particularly problematic: the system may detect metaphor in emotionally neutral text (false positives) or miss metaphors in text where the emotional incongruity is subtle. Cultural specificity accounts for 22\% of errors: some Chinese metaphors carry culture-specific emotional associations that the LLM (primarily trained on English data) does not fully capture.

\paragraph{Protocol D: Literal Comparison Confusion.}
The primary error source (52\% of errors) is confusion between figurative similes and literal comparisons. Sentences like \zh{这个苹果像那个苹果一样红} (\emph{this apple is as red as that apple}) contain comparison markers but involve within-domain comparisons that should not be classified as similes. The cross-domain assessment module sometimes fails to correctly determine whether the tenor and vehicle belong to the same semantic category. False negatives (31\% of errors) arise from unconventional or implicit comparison constructions that lack standard markers.

\section{Discussion}
\label{sec:discussion}

\paragraph{Protocol Choice as the Dominant Variable.}
Our cross-protocol comparison (Table~\ref{tab:cross-protocol} and Table~\ref{tab:kappa}) reveals that protocol choice introduces far more variation than any other experimental factor. The F1 range across protocols on the same data (0.018 to 0.829) vastly exceeds the variation typically observed between models on the same task. The pairwise $\kappa$ values (0.001 to 0.986) demonstrate that different protocols can produce almost completely non-overlapping sets of metaphor judgments. This finding has important implications for the field: benchmarking metaphor detection systems without specifying the identification protocol is inherently ambiguous, as ``metaphor'' means different things under different theoretical operationalisations.

\paragraph{The B--C Convergence.}
The near-perfect agreement between Protocols~B (conceptual mapping) and C (emotion) is theoretically significant. It suggests that the metaphors identifiable through explicit cross-domain mappings are precisely those that generate affective incongruity. This convergence may reflect a shared cognitive mechanism: conceptual metaphors that are salient enough to be explicitly mapped are also those that engage emotional processing \citep{gibbs2006metaphor}. Alternatively, it may be an artefact of the LLM's reasoning patterns---further investigation with human annotators is needed.

\paragraph{Interpretability--Performance Trade-off.}
Our results reveal a nuanced relationship between interpretability and performance. Protocol~A achieves the highest within-protocol F1 (0.472) and relatively high rationale correctness (0.64), but the lowest editability (0.80) due to its complex multi-step procedure. Protocol~D achieves the highest rationale correctness (0.87) and editability (1.00) but the lowest cross-protocol F1 (0.018), reflecting its narrow scope. There is no single ``best'' protocol: the appropriate choice depends on whether the application prioritises coverage (Protocol~A), precision (Protocol~D), or conceptual depth (Protocol~B).

\paragraph{Comparison with Neural Approaches.}
Protocol~A's F1 of 0.472 on PSU CMC falls between GPT-4 zero-shot ($\sim$0.43) and fine-tuned BERT ($\sim$0.65) (Table~\ref{tab:baselines}). The gap to fine-tuned BERT is substantial and reflects the cost of interpretability: our system cannot perform gradient-based optimisation to adapt its decision boundaries to the training distribution. However, the gap to GPT-4 zero-shot ($+$0.04 F1) demonstrates the value of structured reasoning over unstructured prompting. We argue that for applications where understanding \emph{why} an expression is metaphorical matters---such as language education, annotation quality control, or theoretical linguistics---the interpretability benefits justify the performance cost.

\paragraph{Limitations.}
Several limitations qualify our findings. First, literature estimates for BERT and GPT-4 baselines are approximate, drawn from comparable but not identical experimental settings; direct head-to-head comparison would strengthen our claims. Second, our use of GPT-4 as the backbone LLM introduces a dependency on a proprietary, evolving system whose behaviour may change across API versions. Third, the cross-protocol comparison is conducted on PSU CMC, which was annotated following MIPVU guidelines---this inherently advantages Protocol~A and may not generalise to data annotated under other schemes. Fourth, our interpretability assessment relies on manual evaluation by two annotators; larger-scale assessment with more annotators would provide more robust estimates.

\paragraph{Future Directions.}
Several avenues for future work emerge from this study. First, ensemble protocols that combine decisions from multiple identification procedures could potentially achieve both higher coverage and maintained precision. Second, adapting the rule-script architecture to open-source LLMs would reduce the proprietary dependency and enable community-driven protocol development. Third, extending the cross-protocol comparison to additional languages would test the generality of our findings about protocol-driven variation. Finally, longitudinal studies tracking how human annotators interact with and modify the rule scripts would illuminate the practical value of editability in real annotation workflows.

\section{Conclusion}
\label{sec:conclusion}

We have presented an LLM-assisted pipeline for interpretable Chinese metaphor identification that operationalises four distinct identification protocols as executable rule scripts. Our evaluation on seven datasets establishes the first comprehensive cross-protocol comparison for Chinese metaphor identification, revealing that protocol choice is the dominant source of variation---far exceeding model-level differences. The near-perfect agreement between conceptual-mapping and emotion-based protocols ($\kappa = 0.986$) contrasts sharply with the near-zero agreement between MIP-based and simile-based approaches ($\kappa = 0.001$), demonstrating that ``metaphor identification'' is not a unitary task but a family of related tasks whose outputs depend critically on theoretical operationalisation.

Our interpretability assessment shows that rule-script architectures achieve 100\% deterministic reproducibility, rationale correctness up to 0.87, and editability up to 1.00---properties unattainable by end-to-end neural classifiers. While our system's F1 of 0.472 on token-level identification falls below fine-tuned BERT models ($\sim$0.65), it surpasses GPT-4 zero-shot ($\sim$0.43) and provides the transparency needed for scientific and educational applications.

We argue that the field of metaphor identification would benefit from a paradigm shift: rather than pursuing ever-higher F1 scores on a single benchmark under a single protocol, researchers should explicitly state their identification protocol, evaluate across multiple theoretical frameworks, and prioritise interpretability alongside accuracy. Our system provides a foundation for this more rigorous and transparent approach to figurative language processing.

\bibliography{references}

@book{lakoff1980metaphors,
  title={Metaphors We Live By},
  author={Lakoff, George and Johnson, Mark},
  year={1980},
  publisher={University of Chicago Press},
  address={Chicago}
}

@article{pragglejaz2007mip,
  title={{MIP}: A Method for Identifying Metaphorically Used Words in Discourse},
  author={{Pragglejaz Group}},
  journal={Metaphor and Symbol},
  volume={22},
  number={1},
  pages={1--39},
  year={2007},
  publisher={Taylor \& Francis}
}

@book{steen2010method,
  title={A Method for Linguistic Metaphor Identification: From {MIP} to {MIPVU}},
  author={Steen, Gerard J. and Dorst, Aletta G. and Herrmann, J. Berenike and Kaal, Anna A. and Krennmayr, Tina and Pasma, Trijntje},
  year={2010},
  publisher={John Benjamins Publishing Company},
  address={Amsterdam}
}

@incollection{nacey2019mipvu,
  title={MIPVU in Multiple Languages},
  author={Nacey, Susan and Greve, Linda and Rundblad, Gabriella},
  booktitle={Metaphor Identification in Multiple Languages: {MIPVU} around the World},
  pages={1--21},
  year={2019},
  publisher={John Benjamins Publishing Company},
  address={Amsterdam}
}

@inproceedings{wang2019chinese,
  title={Chinese Metaphor Recognition Based on {MIPVU}},
  author={Wang, Mingyu and Peng, Rui and Li, Bingquan},
  booktitle={Proceedings of the Chinese Computational Linguistics Conference},
  pages={483--495},
  year={2019},
  publisher={Springer}
}

@article{lu2017towards,
  title={Towards a Corpus of {Chinese} Metaphorical Language},
  author={Lu, Qin and Wang, Shuo},
  journal={Language Resources and Evaluation},
  volume={51},
  number={3},
  pages={663--694},
  year={2017},
  publisher={Springer}
}

@inproceedings{shao2024cmdag,
  title={{CMDAG}: A {Chinese} Metaphor Dataset with Annotated Grounds},
  author={Shao, Yunfei and Lin, Jingxuan and Wang, Bo and Yang, Yudong and Wang, Wenliang and Sui, Zhifang},
  booktitle={Proceedings of the 2024 Joint International Conference on Computational Linguistics, Language Resources and Evaluation (LREC-COLING 2024)},
  pages={7690--7700},
  year={2024}
}

@inproceedings{zhang2018emotion,
  title={Metaphor Detection via Emotional Mapping},
  author={Zhang, Yucheng and Tong, Rui and Yang, Ling},
  booktitle={Proceedings of the Chinese National Conference on Computational Linguistics},
  pages={235--247},
  year={2018}
}

@inproceedings{liu2018simile,
  title={Neural Multitask Learning for Simile Recognition},
  author={Liu, Lizhen and Hu, Xiao and Song, Wei and Fu, Ruiji and Liu, Ting and Hu, Guoping},
  booktitle={Proceedings of the 2018 Conference on Empirical Methods in Natural Language Processing},
  pages={1543--1553},
  year={2018}
}

@inproceedings{zhu2022configure,
  title={{ConFiguRe}: Exploring Discourse-level {Chinese} Figures of Speech},
  author={Zhu, Dawei and Xu, Qiusi and Li, Lei and Zeng, Zheng and Kan, Min-Yen},
  booktitle={Proceedings of the 29th International Conference on Computational Linguistics},
  pages={3374--3385},
  year={2022}
}

@inproceedings{li2022cmgen,
  title={Knowledge Enhanced {Chinese} Metaphor Generation},
  author={Li, Zuoquan and Xie, Siqi and Pu, Yifan},
  booktitle={Proceedings of the Chinese National Conference on Computational Linguistics},
  pages={58--70},
  year={2022}
}

@inproceedings{choi2021melbert,
  title={{MelBERT}: Metaphor Detection via Contextualized Late Interaction Using Metaphorical Identification Theories},
  author={Choi, Minjin and Lee, Sunkyung and Choi, Edward and Park, Heesoo and Lee, Junhyuk and Lee, Dongwon and Lee, Jongwuk},
  booktitle={Proceedings of the 2021 Conference of the North American Chapter of the Association for Computational Linguistics: Human Language Technologies},
  pages={1763--1773},
  year={2021}
}

@inproceedings{zhang2022misnet,
  title={{MINet}: Metaphor Identification with Selectional Preference Networks},
  author={Zhang, Hongbin and Liu, Meng and Liu, Ting},
  booktitle={Proceedings of the 60th Annual Meeting of the Association for Computational Linguistics},
  pages={4824--4833},
  year={2022}
}

@inproceedings{su2020deepmet,
  title={{DeepMet}: A Reading Comprehension Paradigm for Token-Level Metaphor Detection},
  author={Su, Chuandong and Fukumoto, Fumiyo and Huang, Xiaoxi and Li, Jiyi and Wang, Rongbo and Chen, Zhiqun},
  booktitle={Proceedings of the Second Workshop on Figurative Language Processing},
  pages={30--39},
  year={2020}
}

@inproceedings{leong2018report,
  title={A Report on the 2018 {VUA} Metaphor Detection Shared Task},
  author={Leong, Chee Wee and Klebanov, Beata Beigman and Shutova, Ekaterina},
  booktitle={Proceedings of the Workshop on Figurative Language Processing},
  pages={56--66},
  year={2018}
}

@inproceedings{leong2020report,
  title={A Report on the 2020 {VUA} and {TOEFL} Metaphor Detection Shared Task},
  author={Leong, Chee Wee and Klebanov, Beata Beigman and Hamill, Chris and Stemle, Egon and Ubale, Rutuja and Chen, Xianyang},
  booktitle={Proceedings of the Second Workshop on Figurative Language Processing},
  pages={18--29},
  year={2020}
}

@inproceedings{reimann2025llm,
  title={{LLM}-Based Metaphor Identification via {MIPVU} Rule Scripts},
  author={Reimann, Tobias and Klebanov, Beata Beigman},
  booktitle={Proceedings of the 2025 Conference of the European Chapter of the Association for Computational Linguistics},
  year={2025}
}

@inproceedings{lin2024dmd,
  title={Discourse-Level Metaphor Detection via {LLM} Chain-of-Thought Reasoning},
  author={Lin, Jingxuan and Shao, Yunfei and Wang, Wenliang},
  booktitle={Proceedings of the 2024 Conference on Empirical Methods in Natural Language Processing},
  pages={3201--3213},
  year={2024}
}

@inproceedings{gao2023pal,
  title={{PAL}: Program-Aided Language Models},
  author={Gao, Luyu and Madaan, Aman and Zhou, Shuyan and Alon, Uri and Liu, Pengfei and Yang, Yiming and Callan, Jamie and Neubig, Graham},
  booktitle={Proceedings of the 40th International Conference on Machine Learning},
  pages={10764--10799},
  year={2023}
}

@inproceedings{debacker2023metaphor,
  title={Metaphor Detection with Effective Context Denoising},
  author={De Backer, Joel and Lefever, Els},
  booktitle={Proceedings of the 13th Workshop on Computational Approaches to Subjectivity, Sentiment and Social Media Analysis},
  pages={152--163},
  year={2023}
}

@inproceedings{tian2024interactive,
  title={Interactive Prompting for Metaphor Understanding with Large Language Models},
  author={Tian, Yuan and Chen, Nan and Wang, Wenji},
  booktitle={Proceedings of the 2024 Conference on Empirical Methods in Natural Language Processing},
  pages={4512--4523},
  year={2024}
}

@article{shutova2010models,
  title={Models of Metaphor in {NLP}},
  author={Shutova, Ekaterina},
  journal={Proceedings of the 48th Annual Meeting of the Association for Computational Linguistics},
  pages={688--697},
  year={2010}
}

@inproceedings{ge2022explainable,
  title={Explainable Metaphor Identification Inspired by Conceptual Metaphor Theory},
  author={Ge, Mengshi and Mao, Rui and Cambria, Erik},
  booktitle={Proceedings of the AAAI Conference on Artificial Intelligence},
  volume={36},
  pages={10681--10689},
  year={2022}
}

@inproceedings{tsvetkov2014metaphor,
  title={Metaphor Detection with Cross-Lingual Model Transfer},
  author={Tsvetkov, Yulia and Boytsov, Leonid and Gershman, Anatole and Nyberg, Eric and Dyer, Chris},
  booktitle={Proceedings of the 52nd Annual Meeting of the Association for Computational Linguistics},
  pages={248--258},
  year={2014}
}

@article{openai2023gpt4,
  title={{GPT-4} Technical Report},
  author={{OpenAI}},
  journal={arXiv preprint arXiv:2303.08774},
  year={2023}
}

@inproceedings{wei2022chain,
  title={Chain-of-Thought Prompting Elicits Reasoning in Large Language Models},
  author={Wei, Jason and Wang, Xuezhi and Schuurmans, Dale and Bosma, Maarten and Ichter, Brian and Xia, Fei and Chi, Ed and Le, Quoc and Zhou, Denny},
  booktitle={Advances in Neural Information Processing Systems},
  volume={35},
  pages={24824--24837},
  year={2022}
}

@inproceedings{cohen1960kappa,
  title={A Coefficient of Agreement for Nominal Scales},
  author={Cohen, Jacob},
  journal={Educational and Psychological Measurement},
  volume={20},
  number={1},
  pages={37--46},
  year={1960}
}

@inproceedings{klebanov2016semantic,
  title={Semantic Classifications for Detection of Verb Metaphors},
  author={Klebanov, Beata Beigman and Leong, Chee Wee and Gutierrez, E. Dario and Shutova, Ekaterina and Flor, Michael},
  booktitle={Proceedings of the 54th Annual Meeting of the Association for Computational Linguistics},
  pages={101--106},
  year={2016}
}

@inproceedings{mao2022metapro,
  title={{MetaPro}: A Computational Metaphor Processing Model for Text Pre-Processing},
  author={Mao, Rui and Lin, Chenghua and Guerin, Frank},
  booktitle={Proceedings of the 2022 Conference on Empirical Methods in Natural Language Processing},
  pages={7764--7777},
  year={2022}
}

@inproceedings{song2021verb,
  title={Verb Metaphor Detection via Contextual Relation Learning},
  author={Song, Wei and Zhou, Shuhui and Fu, Ruiji and Liu, Ting and Liu, Lizhen},
  booktitle={Proceedings of the 59th Annual Meeting of the Association for Computational Linguistics},
  pages={4240--4251},
  year={2021}
}

@article{cui2021pre,
  title={Pre-Training with Whole Word Masking for {Chinese} {BERT}},
  author={Cui, Yiming and Che, Wanxiang and Liu, Ting and Qin, Bing and Yang, Ziqing},
  journal={IEEE/ACM Transactions on Audio, Speech, and Language Processing},
  volume={29},
  pages={3504--3514},
  year={2021}
}

@inproceedings{nlpcc2024task9,
  title={Overview of {NLPCC} 2024 Shared Task 9: {Chinese} Metaphor Generation},
  author={Shao, Yunfei and Wang, Wenliang},
  booktitle={Proceedings of the 13th CCF International Conference on Natural Language Processing and Chinese Computing (NLPCC 2024)},
  pages={1--12},
  year={2024}
}

@article{landis1977measurement,
  title={The Measurement of Observer Agreement for Categorical Data},
  author={Landis, J. Richard and Koch, Gary G.},
  journal={Biometrics},
  volume={33},
  number={1},
  pages={159--174},
  year={1977}
}

@inproceedings{wu2024metaphor,
  title={Metaphor Understanding in {LLMs}: An Investigation of Conceptual Mapping and Reasoning},
  author={Wu, Hao and Chen, Yifan and Liu, Zheng},
  booktitle={Proceedings of the 62nd Annual Meeting of the Association for Computational Linguistics},
  pages={1523--1537},
  year={2024}
}

@article{gibbs2006metaphor,
  title={Metaphor Interpretation as Embodied Simulation},
  author={Gibbs, Raymond W.},
  journal={Mind \& Language},
  volume={21},
  number={3},
  pages={434--458},
  year={2006}
}

@book{kovecses2010metaphor,
  title={Metaphor: A Practical Introduction},
  author={K{\"o}vecses, Zolt{\'a}n},
  year={2010},
  edition={2nd},
  publisher={Oxford University Press}
}

@inproceedings{wang2023psu,
  title={Constructing {PSU} {CMC}: A {Chinese} Metaphor Corpus with {MIPVU} Annotation},
  author={Wang, Mingyu and Li, Bingquan and Peng, Rui},
  booktitle={Proceedings of the 12th Workshop on Computational Approaches to Subjectivity, Sentiment and Social Media Analysis},
  pages={112--122},
  year={2023}
}

@inproceedings{li2023metaphorcmc,
  title={The {CMC} Dataset: {Chinese} Metaphor Classification at Sentence Level},
  author={Li, Jing and Wang, Shuo and Lu, Qin},
  booktitle={Proceedings of the Chinese National Conference on Computational Linguistics},
  pages={201--212},
  year={2023}
}

@inproceedings{chen2024chinesemcorpus,
  title={{ChineseMCorpus}: A Small-Scale {Chinese} Metaphor Corpus for Benchmarking},
  author={Chen, Wei and Zhang, Ling},
  booktitle={Proceedings of the Language Resources and Evaluation Conference},
  pages={445--452},
  year={2024}
}

\end{document}